\documentclass[sigconf]{aamas} 

\usepackage{balance} 
\usepackage{hyperref}
\usepackage{url}
\usepackage{graphicx}
\usepackage{amsmath}
\DeclareMathOperator*{\argmax}{arg\,max}

\usepackage{algorithm}
\usepackage{algorithmicx}
\usepackage{algpseudocode}
\usepackage{diagbox}

\setcopyright{ifaamas}
\acmConference[AAMAS '24]{Proc.\@ of the 23rd International Conference
on Autonomous Agents and Multiagent Systems (AAMAS 2024)}{May 6 -- 10, 2024}
{Auckland, New Zealand}{N.~Alechina, V.~Dignum, M.~Dastani, J.S.~Sichman (eds.)}
\copyrightyear{2024}
\acmYear{2024}
\acmDOI{}
\acmPrice{}
\acmISBN{}

\acmSubmissionID{86}

\title[AAMAS-2024 Formatting Instructions]{Safe Reinforcement Learning with Free-form Natural Language Constraints and Pre-Trained Language Models}

\author{Xingzhou Lou}
\affiliation{
  \institution{Institute of Automation, CAS}
  \city{Beijing}
  \country{China}}
\email{louxingzhou2020@ia.ac.cn}
\authornote{Also with School of Artificial Intelligence, University of Chinese Academy of Sciences}

\author{Junge Zhang}
\affiliation{
  \institution{Institute of Automation, CAS}
  \city{Beijing}
  \country{China}}
\email{jgzhang@nlpr.ia.ac.cn}
\authornotemark[1]
\authornote{Correspondence.}

\author{Ziyan Wang}
\affiliation{
  \institution{King's College London}
  \city{London}
  \country{the United Kingdom}}
\email{ziyan.wang@kcl.ac.uk}

\author{Kaiqi Huang}
\affiliation{
  \institution{Institute of Automation, CAS}
  \city{Beijing}
  \country{China}}
\email{kqhuang@nlpr.ia.ac.cn}
\authornotemark[1]

\author{Yali Du}
\affiliation{
  \institution{King's College London}
  \city{London}
  \country{the United Kingdom}}
\email{yali.du@kcl.ac.uk}
\authornotemark[2]

\begin{abstract}
Safe reinforcement learning (RL) agents accomplish given tasks while adhering to specific constraints. Employing constraints expressed via easily-understandable human language offers considerable potential for real-world applications due to its accessibility and non-reliance on domain expertise. Previous safe RL methods with natural language constraints typically adopt a recurrent neural network, which leads to limited capabilities when dealing with various forms of human language input. Furthermore, these methods often require a ground-truth cost function, necessitating domain expertise for the conversion of language constraints into a well-defined cost function that determines constraint violation. To address these issues, we proposes to use pre-trained language models (LM) to facilitate RL agents' comprehension of natural language constraints and allow them to infer costs for safe policy learning. Through the use of pre-trained LMs and the elimination of the need for a ground-truth cost, our method enhances safe policy learning under a diverse set of human-derived free-form natural language constraints. Experiments on grid-world navigation and robot control show that the proposed method can achieve strong performance while adhering to given constraints. The usage of pre-trained LMs allows our method to comprehend complicated constraints and learn safe policies without the need for ground-truth cost at any stage of training or evaluation. Extensive ablation studies are conducted to demonstrate the efficacy of each part of our method.
\end{abstract}


\newcommand{\BibTeX}{\rm B\kern-.05em{\sc i\kern-.025em b}\kern-.08em\TeX}

\makeatletter
\gdef\@copyrightpermission{
	\begin{minipage}{0.3\columnwidth}
		\href{https://creativecommons.org/licenses/by/4.0/}{\includegraphics[width=0.90\textwidth]{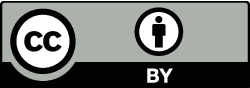}}
	\end{minipage}\hfill
	\begin{minipage}{0.7\columnwidth}
		\href{https://creativecommons.org/licenses/by/4.0/}{This work is licensed under a Creative Commons Attribution International 4.0 License.}
	\end{minipage}
	\vspace{5pt}
}
\makeatother

\begin{document}


\pagestyle{fancy}
\fancyhead{}


\maketitle 

\section{Introduction}\begin{figure*}[ht]
    \centering
    \includegraphics[width=.9\textwidth]{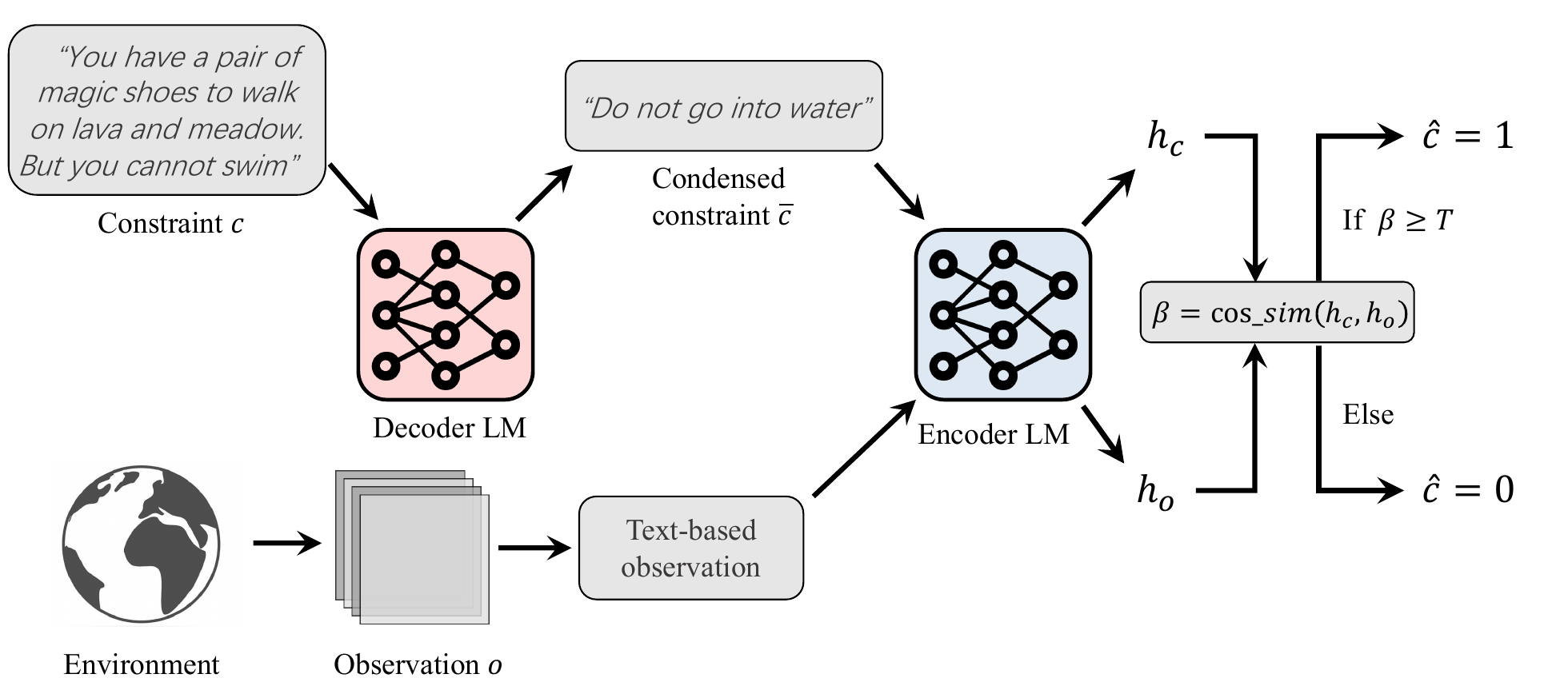}
    \caption{Cost prediction in the proposed method. The decoder LM condenses the semantic meaning of the constraint to eliminate ambiguity and redundancy. The encoder LM encodes the condensed constraints and text-based observations into embeddings according to their semantic meaning. If cosine similarity between the embeddings is greater than threshold $T$, the model will predict the constraint is violated (predicted cost $\hat{c}=1$), otherwise $\hat{c}=0$. Embedding $h_c$ is also used later as input to condition the policy network.}
    \label{fig:struc}
\end{figure*}

In recent years,  great success has been achieved with reinforcement learning (RL) algorithms on multiple domains such as robotic control \cite{kaufmann2023champion}, resource allocation \cite{zheng2022ai} and games \cite{vinyals2019grandmaster,perolat2022mastering,du2019liir}. In real-world settings such as autonomous driving \cite{gu2022review}, RL agents are not given complete freedom for safety, fairness or other concerns and must adhere to specific constraints provided by humans. To this end, safe RL is widely studied, following the constrained Markov Decision Process (CMDP) \cite{altman1999constrained} framework, where agents must satisfy the constraints given by auxiliary costs when optimizing their policies to maximize the objective function. 

To learn a safe policy, safe RL algorithms \cite{achiam2017constrained,ray2019benchmarking,yang2020projection,yang2022constrained,gu2023safe} utilize ground-truth cost value given by a well-defined cost function in CMDP. Because the conditions for violating constraints are different, cost functions need to be defined for all constraints case by case and require task-specific domain knowledge, which can be very expensive and inefficient. For example, in a navigation task, the agent should not get close to objects that may cause damage to it. To define such a cost function, one must be aware of \emph{all} the harmful objects in the environment, otherwise the agent may be harmed by some unexpected hazards. This severely limits safe RL's application to general and complex tasks where transforming constraints into cost functions can be expensive and requires specific domain knowledge. 

One intuitive and easily-accessible approach to regulate safe RL agents is to provide constraints via human language \cite{kaplan2017beating,prakash2020guiding,yang2021safe,kazantzidis2022train}. By providing natural language constraints, potential end users can easily regularize and interact with agents. However, without a ground-truth cost function, it is difficult to determine whether natural language constraints are violated. To avoid using ground-truth costs, previous method \cite{yang2021safe} learns to identify entities relevant to the natural language constraints in the environment so that agents can predict whether the constraints are violated. However, it requires additional computation to model the entities and may result in inaccurate predictions in complex environments. And since the model is trained in a supervised fashion, it is incapable of handling free-form or new natural language constraints from humans.

In this paper, to deal with the problem of constraint violation prediction in safe RL with natural language constraints, we propose to adopt pre-trained language models (LM) \cite{devlin2018bert,brown2020language} to comprehend and encode the constraints and predict costs for policy learning. 
Specifically, a decoder-based LM (such as GPT \cite{brown2020language}) is adopted to eliminate ambiguity and redundancy and extract semantic meaning from the constraints. As decoder LMs are pre-trained on a very large corpus of data and align with human values, they are able to generate high-quality constraints with condensed semantic meaning if we prompt them properly. To condition the policy which is parameterized by neural networks with natural language constraints, an encoder-based LM (such as BERT \cite{devlin2018bert}) encodes constraints into embeddings according to their semantic meaning. We also use the encoder LM to encode the text-based observations. By computing the semantic similarities between constraint embeddings and description embeddings, we can determine whether the constraint is violated. To distinct constraint embeddings from each other, we adopt a contrastive loss to fine-tune the encoder LM so that constraints have similar embeddings only when they have similar semantic meaning.

We empirically validate the effectiveness of our method in two environments: Hazard-World-Grid \cite{chevalier2018minimalistic} and SafetyGoal from Safety-Gymnasium \cite{Safety-Gymnasium}. Hazard-World-Grid is a grid-world navigation task, and SafetyGoal is aimed at safe robot control. In both environments, agents must navigate the world to achieve the goal and avoid potential hazards according to the constraints given by free-form natural language constraints from humans. Compared to constraints given by structured language in previous works \cite{prakash2020guiding,yang2021safe}, pre-trained LMs' strong capabilities in natural language processing allow our method to deal with much more complicated forms of constraints. Our method achieves strong performance regarding both reward and cost without the need for ground-truth costs at any stage of training or evaluation. Experimental results demonstrate that the proposed method is able to constrain the agents effectively. Extensive ablation studies show the decoder LM and encoder LM are both necessary in order to predict costs accurately. 

Contributions of this paper are three-fold:
\begin{enumerate}
    \item We introduce pre-trained LMs to safe RL with natural language constraints to replace the recurrent neural network in previous works, which can only handle structured or fixed-form natural language constraints.
    \item An encoder LM and a decoder LM are adopted for accurate cost prediction, which is essential for safe policy learning. The LM-based cost prediction enables agents to learn safe policy without the need for ground-truth cost at any stage of training.
    \item Experiments on grid-world navigation and robot control both show the proposed method can make agent learn safe policies to follow the constraints. Extensive ablation studies confirm the efficacy of our LM-based cost prediction module.
\end{enumerate}
\section{Related Work}
\subsection{Safe Reinforcement Learning}
Safe RL predominantly aims to train policies that maximize rewards while obeying specific constraints\cite{garcia2015comprehensive,gu2022review}. Safety is a significant practical problem in RL's real-world applications, especially in safety-critical scenarios such as autonomous driving \cite{kiran2021deep} and power allocation \cite{nasir2019multi}. The most straightforward safe RL algorithm is PPO-Lagrangian \cite{ray2019benchmarking}, which uses adaptive penalty coefficients to enforce constraints with a Lagrangian multiplier to the reward sum objective and solve the problem by solving the equivalent max-min optimization method. Constrained Policy Optimization (CPO) \cite{achiam2017constrained} is a policy search algorithm that derives policy improvement steps that guarantee both an increase in reward and satisfaction of constraints on costs. Projection-Based Constrained Policy Optimization (PCPO) \cite{yang2020projection} first performs local reward improvement update and then projects the updated policy to the constraint set to make sure the updated policy is always safe. However, typical safe RL algorithms require ground-truth costs to do policy updates, which are often unavailable in tasks where constraints are only given by natural language.

\subsection{Reinforcement Learning with Natural Language} 
Multiple previous studies have combined reinforcement learning with natural language. RL is widely used in text-based tasks such as machine translation \cite{wu2018study} and text games \cite{cote2019textworld} to achieve strong performance. Similarly, He et al. \cite{he2015deep} studied RL agents with a natural language action space. Natural language can be used to instruct or inform agents to get higher rewards. Kaplan et al. \cite{kaplan2017beating} used natural language instructions from an expert to aid RL agents in accomplishing the tasks. Agents receive additional rewards if they follow the expert's natural language instructions. Goyal et al. \cite{goyal2019using} proposed the use of natural language instructions to perform reward shaping to improve sample efficiency of RL algorithms. Previous works also used natural language to constrain agents to behave safely. Prakash et al. \cite{prakash2020guiding} trained a constraint checker in a supervised fashion to predict whether the natural language constraints are violated and guide RL agents to learn safe policies. However, they required ground-truth cost to train the constraint checker, which may be unavailable in many tasks. Yang et al .\cite{yang2021safe} trained a constraint interpreter to predict which entities in the environment may be relevant to the constraint and used the interpreter to predict costs. Although their method did not require ground-truth cost, the interpreter had to model and predict all entities in the environment, which required additional computation and may lead to inaccurate results in complex tasks. For comparison, our method does not need ground-truth costs and adopts pre-trained LMs to predict constraint violations, which avoids training extra modules and can harness the knowledge in large pre-trained LMs.

\subsection{Pre-Trained Language Models} 
By utilizing the power of Transformer \cite{vaswani2017attention}, pre-trained LMs have drawn enormous attention in recent years. Based on the encoder part of Transformer, encoder-based LMs such as Bidirectional Encoder Representations from Transformers (BERT) \cite{devlin2018bert} extract semantic meaning and learn representations for text inputs by joint conditioning on their context and can be easily fine-tuned for downstream tasks. Based on the decoder part of Transformer, decoder-based LMs such as Generative Pre-trained Transformer (GPT) \cite{gpt4} and Llama \cite{touvron2023llama} are trained to generate texts given previous contexts, and thus good at text generation tasks. Both encoder-based and decoder-based LMs are pre-trained on a very large corpus of text and thus possess strong natural language processing ability. Previous studies have tried to introduce pre-trained LMs to RL agents. Du et al. \cite{du2023guiding} utilized decoder LMs to generate novel goals for agents and encoder LMs to provide extrinsic rewards to encourage agents to achieve the generated goal and explore the environment. Hu et al. \cite{hu2023language} used pre-trained LMs to generate policies conditioned on human instructions and help RL agents converge to equilibria aligned with human preferences, which significantly boosted their human-AI coordination performance. Nottingham et al. \cite{nottingham2023embodied} proposed to utilize pre-trained LMs to hypothesize a world model to guide RL agents' exploration and improve sample efficiency. The hypothesized world model is also verified by world experience of RL agents from interactions. But to the best of our knowledge, our work is the first to apply pre-trained LMs to the field of safe RL.

\section{Preliminaries}
Safe RL mostly models problems as Constrained Markov Decision Process (CMDP) \cite{altman1999constrained}. CMDP consists of a tuple $<S,A,T,R,\gamma,C>$, where $S$ is the state space, $A$ is the action space, $T$ is the state transition function, $R$ is the reward function, $C$ is some given cost function and $\gamma$ is the discount factor.  The objective for the agent is to maximize accumulated reward $J=\mathbb{E}_\pi[\sum\limits_{t=0}\gamma^tR_t]$ and minimize the accumulated cost $J_C=\mathbb{E}_\pi[\sum\limits_{t=0}\gamma^tc_t]$ to satisfy the constraints at the same time. The learning objective is given by
\begin{equation}\label{cmdp_obj}
    \begin{aligned}
        \max\limits_\pi J(\pi)&=\mathbb{E}_\pi\left[\sum\limits_{t=0}\gamma^tR_t\right]\ \  
        s.t.\ \ J_C(\pi)&=\mathbb{E}_\pi\left[\sum\limits_{t=0}\gamma^tc_t\right]\leq H,
    \end{aligned}
\end{equation}
where $H$ is the constraint violation budget.

However, we consider the problem where the cost function $C$ is not known but instead implied by some free-form natural language. Thus, instead of CMDP, we model the problem by another tuple $<S,A,T,R,\gamma,M,X>$, which is an MDP augmented by a constraint transformation function $M$ and natural language constraint space $X$. $M:X\rightarrow C^x$ maps some natural language constraint $x\in X$ to a cost function $C^x$, where $C^x:S\times A\rightarrow \{0,1\}$ decides whether the agent has violated the natural language constraints. The agent only knows the constraint $x$ and has no knowledge of ground-truth cost $C^x(s_t,a_t)$ under any circumstances. The constraint $x$ is sampled at the beginning of each episode and remains consistent within the episode. As each natural language constraint $x$ corresponds to a cost function $C^x$, we will refer natural language constraints with variable $c$ in what follows.
\section{Methodology}
In this section, we introduce our proposed method. We first introduce our LM-based cost prediction module and then show how the policy is trained with the predicted costs. 
\subsection{Cost Prediction Module}\label{sec:cost_pred}
We give the structure of the cost prediction module in Fig. \ref{fig:struc}. 
In general, to predict whether the constraint is violated, we adopt a decoder LM (GPT) to eliminate ambiguity and condense semantic meaning in free-form natural language constraint $c$ and an encoder LM (BERT) to extract semantic information from the condensed natural language constraint $\overline{c}$ and transform the constraints into embedding $h_c$. Condensed constraint $\overline{c}$ aligns with the original constraint $c$ on which entity or behaviour the agents are prohibited from. This is necessary as free-form constraint $c$ from humans may be verbose or semantically vague, which will severely affect the cost prediction result. Since decoder LM GPT aligns with human values \cite{ziegler2019fine}, it can eliminate ambiguity within the human natural language constraints and better summarize the constraints. We empirically show in the experiment section that without the decoder LM, the performance of the cost prediction module will drop significantly. The prompt we use for the decoder LM is provided in the supplementary material.

To ensure that constraints with the same semantic meaning have similar embeddings, we use a contrastive loss to fine-tune the encoder LM. For a batch of natural language constraint pairs, the contrastive loss is given by
\begin{equation}\label{eq:contrast}
    \mathcal{L}_c=\frac{1}{n}\sum\limits_{i=1}^n\left[\frac{1}{2}\left(Y^i-{cosine\_sim}(h_{c_1}^i,h_{c_2}^i)\right)^2\right],
\end{equation}
where $h_{c_1}^i,h_{c_2}^i$ are embeddings of $i$th natural language constraint pair $(c^i_1,c^i_2)$. The constraints are first processed by the decoder LM and then encoded by the encoder LM. Target $Y^i=1$ when constraints $c_1^i,c_2^i$ prohibit the agent from the same entities or behaviours, otherwise $Y^i=0$. Cosine similarity $cosine\_sim(h_{c_1},h_{c_2})=\frac{h_{c_1}\cdot h_{c_2}}{\|h_{c_1}\|\|h_{c_2}\|}$.

The contrastive loss enables the encoder LM to recognise constraints' semantic similarity \cite{reimers-2019-sentence-bert}. As a result, constraints concerning the same entities and behaviours will have embeddings with high cosine similarity and vice versa.

The constraint embedding $h_c$ is later used together with the observation to determine whether the constraint is violated. As $h_c$ is semantic embedding, we have to pre-process observation $o$ into text-based observation, so that it can be used for cost prediction with the semantic embedding of constraints. Text-based observations are widely used as they provide easily accessible information for LMs to process \cite{yao2022react,zhang2023building,wang2023voyager,du2023guiding}. The transformation can be done via template matching \cite{brunelli2009template} or caption models with supervised learning \cite{hossain2019comprehensive}. In our experiments, a descriptor will automatically analyze observation-action pairs and give natural language descriptions based on pre-defined templates. After obtaining the text-based observation, the encoder LM will encode it into observation embedding $h_o$, whose semantic meaning contains information about the current situation of the agent. Finally, the predicted cost $\hat{c}$ is given by
\begin{equation}\label{eq:c_predict}
        \hat{c}=\left\{
    \begin{aligned}
        & 1 \text{\ \ \ if\ \ \ }cosine\_sim(h_c,h_o)>T \\
        & 0\text{\ \ \ otherwise }
    \end{aligned}
    \right. ,
\end{equation}
where $T$ is a hyperparameter defining the threshold and sensitivity of the cost prediction module, $h_c$ is the constraint embedding, $h_o$ is the observation embedding, and $cosine\_sim$ is cosine similarity. $\hat{c}=1$ means that the cost prediction module predicts the constraint has been violated. The insight in devising such a cost function is that If the constraint and observation are semantically similar to each other, it is very likely that the agent has taken undesired actions or encountered undesired situations, resulting in an observation similar to the constraints. 
\subsection{Policy Training}\begin{algorithm}[t]
        \caption{PPO with LM-based Cost Prediction (PPO-CP)}
        \begin{algorithmic}[1]
            \State Initialize value function network $\phi$, cost value function network $\phi_c$, policy network $\theta$, decoder LM $M_d$ and encoder LM $M_e$, Lagrange Multiplier update stepsize $\eta$
            \For{each episode}
                \State Sample a natural language constraint $x$ 
                \State Condense and extract the semantic meaning of x with $M_d$
                \State Encode the condensed constraint with $M_e$ to get constraint embedding $h_c$
                \State Rollout the policy with constraint $h_c$ and get trajectory $\{o_t,a_t,o'_t,r_t\}_{t=1,..,n}$
                \For{t \textbf{in} \{1,..,n\}}
                    \State Transform $o_t$ into text-based observation and encode it with $M_e$ to get observation embedding $h_o$
                    \State Predict cost $\hat{c}_t$ according to Eq. \ref{eq:c_predict}
                \EndFor
                \State Calculate loss $L^{VF}$ and $L^{CVF}$ for value function $\phi$ and cost value function $\phi_c$
                \State Calculate policy loss $L^{CLIP}$ and $L_C^{CLIP}$ with value function $\phi$ and cost value function $\phi_c$
                \State Update value function $\phi$, cost value function $\phi_c$ and policy network $\theta$ according to Eq. \ref{eq:total}
                \State Update $\alpha$ by stepsize $\eta$ to maximize $L_\alpha$ in Eq. \ref{eq:alpha}.
            \EndFor
        \end{algorithmic}
\end{algorithm}
With the predicted cost $\hat{c}$, we are ready to train our safe RL agents. It is worth noting that our method does not require the ground-truth cost under any circumstances during training or evaluation, which is distinct compared to other existing safe RL algorithms. We integrate the cost prediction module to proximal policy optimization (PPO) \cite{schulman2017proximal} with Lagrange multiplier \cite{ray2019benchmarking}, so that the agents can maximize rewards while adhering to specific constraints at the same time. 

We define the cost return, cost value function and cost advantage function as $J_C, V^\pi_C$ and $A^\pi_C$ in analogy to return $J$, value function $V^\pi$ and advantage function $A^\pi$.

The policy $\pi$ is obtained by
$$\pi=\argmax\limits_\pi J(\pi)-\alpha J_{C}(\pi),$$
where $J(\pi)=\mathbb{E}_\pi\left[\sum\limits_{t=0}^T\gamma^tr_t\right]$ is the expected reward sum under policy $\pi$ and $J_{C}(\pi)=\mathbb{E}_\pi\left[\sum\limits_{t=0}^T\gamma^t\hat{c}_t\right]$ is the expected cost sum, where $\hat{c}_t$ are predicted cost at timestep $t$, and $\alpha$ is the Lagrange multiplier.\begin{figure*}[ht]
    \centering
    \includegraphics[width=\textwidth]{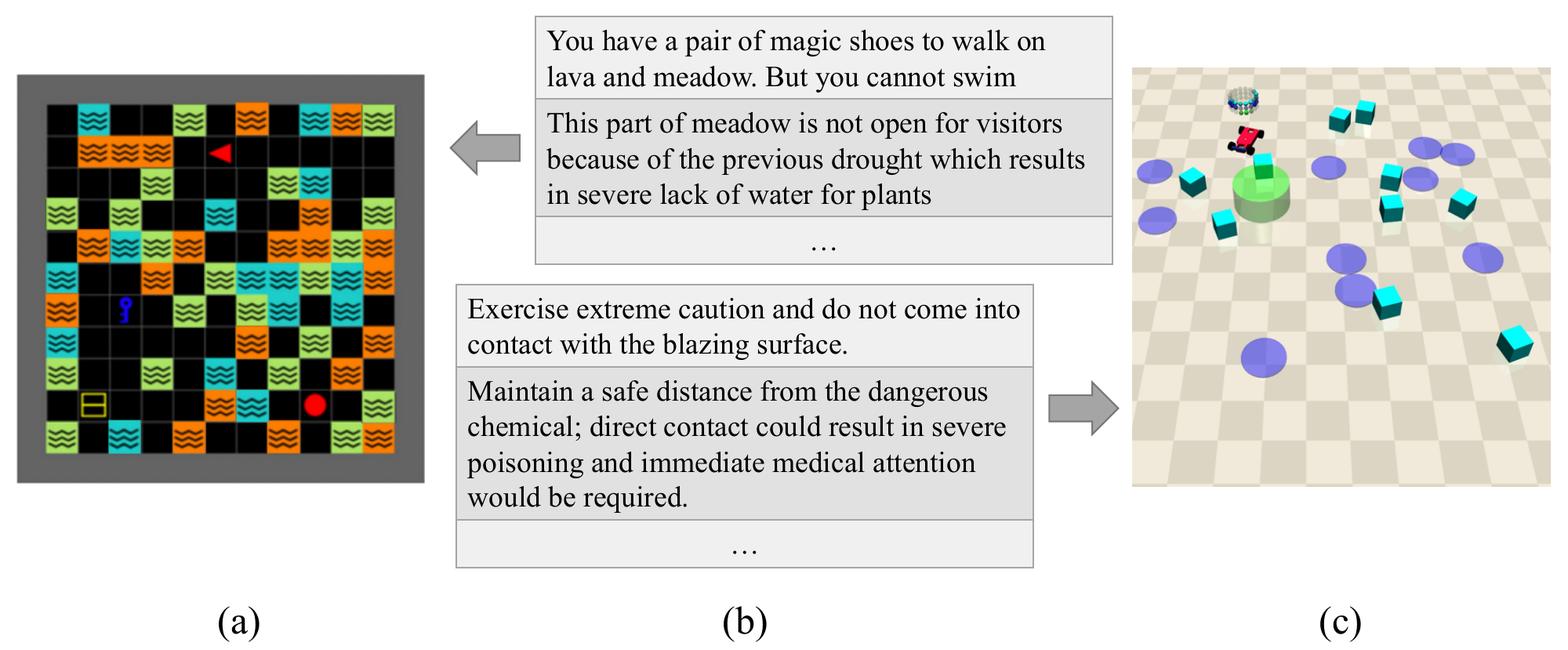}
    \caption{(a) One layout in Hazard-World-Grid, where orange tiles are lava, blue tiles are water and green tiles are grass. (c) Robot navigation task SafetyGoal built-in Safety-Gymnasium \cite{Safety-Gymnasium}, where there are multiple types of objects in the environment. In both environments (a) and (c), agents have to reach goals while avoiding some type of terrain or objects specified in the natural language constraints. (b) Constraint examples for two environments in our experiments. Compared to constraints by structured language in previous works, constraints in our experiments are much more free-form and less intuitive.}
    \label{fig:envs}
\end{figure*}

The training of value function $V^\pi$ and cost value function $V_C^\pi$ is updated by minimizing the corresponding mean squared TD-error
\begin{align}
    L^{VF}&=\mathbb{E}_\pi\left[\left(r^t_{(i)}+\gamma V(s_{(i)}^{t+1})-V(s_{(i)}^t)\right)^2\right],\\
    L^{CVF}&=\mathbb{E}_\pi\left[\frac{1}{2}\left(\hat{c}^t_{(i)}+\gamma V_C(s_{(i)}^{t+1},h_c)-V_C(s_{(i)}^t,h_c)\right)^2\right],
\end{align}
where $h_c$ is the constraint embedding from the encoder LM, $n$ is batch size and $\hat{c}^t$ is the predicted cost.

To maximize return $J$ and minimize cost return $J_C$, two clipped policy losses are adopted:
\begin{equation}
    \begin{aligned}
        L^{CLIP}=\mathbb{E}_t\Bigg[\min&\Bigg(\frac{\pi(a^t|s^t,h_c)}{\pi_{old}(a^t|s^t,h_c)}A^t,\\
        &clip\Big(\frac{\pi(a^t|s^t,h_c)}{\pi_{old}(a^t|s^t,h_c)},1-\epsilon,1+\epsilon\Big)A^t\Bigg)\Bigg],
    \end{aligned}
\end{equation}
\begin{equation}
    \begin{aligned}
        L^{CLIP}_C=\mathbb{E}_t\Bigg[\min&\Bigg(\frac{\pi(a^t|s^t,h_c)}{\pi_{old}(a^t|s^t,h_c)}A_C^t,\\
        &clip\Big(\frac{\pi(a^t|s^t,h_c)}{\pi_{old}(a^t|s^t,h_c)},1-\epsilon,1+\epsilon\Big)A_C^t\Bigg)\Bigg],
    \end{aligned}
\end{equation}
where advantage $A^t$ is obtained by generalized advantage estimator \cite{schulman2015high} $A^t=\sum\limits_{l=0}^{\infty}(\gamma\lambda)^l\delta_V^{t+l}$, $\delta^V_t=r+\gamma V(s^{t+1})-V(s^t)$ is the TD-error. $A^t_C$ is obtained by replacing $V$ with cost value function $V_C$.

Together, the value networks and policy network is trained by maximizing the total loss
\begin{equation}
\label{eq:total}
    L=L^{CLIP}-\alpha L^{CLIP}_C-c_1L^{VF}-c_2L^{CVF},
\end{equation}
where $c_1$ and $c_2$ are coefficients for value network updates. 

Lagrange multiplier $\alpha$ is updated after each training iteration. Loss for updating $\alpha$ is given by
\begin{equation}
\label{eq:alpha}
    L_\alpha=\alpha(\mathbb{E}_\pi[\sum_t\hat{c}_t]-H),
\end{equation}
where $H$ is the hyperparameter of cost budget from Eq. \ref{cmdp_obj}. By updating $\alpha$ with a small stepsize to maximize $L_\alpha$, penalty coefficient $\alpha$ will increase if the expected cost sum is larger than $H$, and otherwise decrease (but will always be non-negative). This will enable agents to learn policies that satisfy the constraint quickly and achieve rewards as high as possible.

By doing the update in Eq. \ref{eq:total} iteratively, the agent will learn to maximize the reward and minimize constraint violations simultaneously. The agent will learn a safe policy and accomplish the given task while trying to follow the natural language constraints. The pseudo-code of the proposed method is given in Algorithm 1.
\section{Experiment}
\subsection{Experiment Settings}\begin{figure*}[ht]
    \centering
    \includegraphics[width=\textwidth]{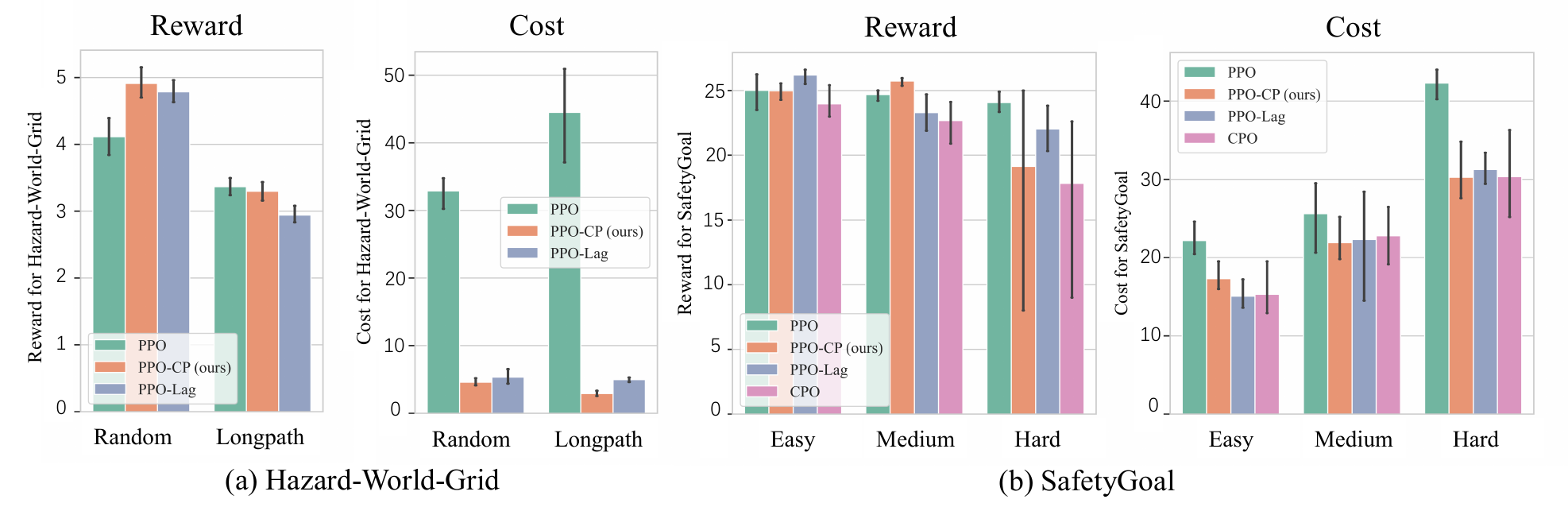}
    \caption{Experiment results on Hazard-World-Grid and SafetyGoal. \emph{random} and \emph{longpath} are two layouts in Hazard-World-Grid. \emph{Easy}, \emph{Medium} and \emph{Hard} are three levels in SafetyGoal. There are two objects of each hazard type in \emph{Easy}, four in \emph{Medium} and six in \emph{Hard}. Thus, in total, there are 8 hazard objects in \emph{Easy}, 16 in \emph{Medium} and 24 in \emph{Hard}. PPO-Lag refers to the baseline method PPO-Lagrangian, and PPO-CP refers to the proposed method, where the suffix CP means cost prediction. From the results, the proposed method successfully learns a safer policy with the predicted cost. CPO is proposed for robot locomotion tasks. Thus, we do not include it in the first two tasks from Hazard-World-Grid. It is worth noting that in some tasks, the proposed method learns safer policies than the baselines using ground-truth costs. This is because the cost prediction module may mistakenly think of some safe but risky actions as violating the constraints, resulting in a more conservative policy.}
    \label{fig:main_res}
\end{figure*}
We evaluate the proposed method on two tasks: Hazard-World-Grid \cite{yang2021safe} and SafetyGoal \cite{Safety-Gymnasium} shown in Fig. \ref{fig:envs}. Each task is accompanied by 20 different constraints to regularize the agents. A full list of detailed constraints for each environment is given in the supplementary material. The pre-trained decoder LM and encoder LM we use are \emph{gpt-3.5-turbo} \cite{brown2020language} and \emph{all-MiniLM-L12-v2} from sentence BERT \cite{reimers2019sentence}. So, in this section, we also refer to the encoder LM as BERT and the decoder LM as GPT. The detailed constraints, prompts for the decoder LMs to condense constraints and how we transform observation into text-based observations are provided in the supplementary material. On each task, all methods are run for 4 times with different random seeds. The baseline methods we compare here are PPO \cite{schulman2017proximal}, PPO-Lagrangian (PPO-Lag) \cite{ray2019benchmarking} and CPO \cite{achiam2017constrained}. PPO only optimizes policies to maximize reward and does not consider the constraints. PPO-Lag enforces constraints with a Lagrangian multiplier to the reward objective and optimizes the policy by solving the equivalent min-max problem. CPO derives policy improvement steps that guarantee both an increase in reward and satisfaction of constraints on costs and is a strong baseline for robot locomotion tasks with constraints. For these two methods, we input the natural language constraints encoded by our encoder LM to the policy network and provide them ground-truth costs to do policy learning, while our proposed method it does not have access to the ground-truth costs at all. Implementation and hyperparameters of the baseline methods can be found here \footnote{\url{https://github.com/PKU-Alignment/Safe-Policy-Optimization}}. Our method keeps the default hyperparameter and network details of PPO-Lag but use the predicted costs to replace the ground-truth costs.

\noindent\textbf{Hazard-World-Grid} In Hazard-World-Grid, agents navigate in a grid-world to find goal objects $\{ball,box,key\}$. Reward for reaching $ball$ is 1, 2 for $box$ and 3 for $key$. The reward will linearly decay to $0.1$ times its original value w.r.t. timesteps within an episode, which means the sooner an object is found, the higher the reward. $\{lava,water,grass\}$ are hazards in the grid-world. In each episode, the agent will be told to avoid a specific kind of hazard by a natural language constraint. At the end of an episode, the ground-truth cost will be the total number of times that the constraint is violated. An episode terminates when all three goal objects are found, or maximum timesteps (300 in our experiments) are reached. Two layouts in this environment are used in our experiment: (1) \emph{random}, where lava, water and grass tiles are randomly scattered in the grid-world as in Fig. \ref{fig:envs} (a); (2) \emph{longpath}, where the grid-world is filled with lava, except for a single safe path with many turns and no hazards on it. The difficulty of \emph{longpath} is that the agent can only walk on the path if the constraint is to avoid lava. while in any other case, it can behave freely. Threshold $T$ for cost prediction is $0.4$ in this environment.

\noindent\textbf{SafetyGoal} In SafetyGoal, agents control a robot to navigate a 2D plane to reach the Goal's location while circumventing a given type of object. The objects are categorized into \{poisonous-hazard,burning-hazard,radioactive-hazard,bio-hazard\}. The goal will be randomly relocated when the robot reaches its location. Similar to Hazard-World-Grid, at the beginning of each episode, a natural language constraint will be given to tell the agent which type of object to avoid. The cost will be the total number of constraint violations within the episode. An episode terminates when maximum timesteps are reached (1000 in our experiments). According to the number of hazard objects of each type in the environment, we name the tasks \emph{Easy}, \emph{Medium} and \emph{Hard}, where in \emph{Easy}, there are 8 hazard objects in total, 16 in \emph{Medium} and 24 in \emph{Hard}. Threshold $T$ for cost prediction is $0.55$.
\subsection{General Results}\begin{figure*}[t]
    \centering
    \includegraphics[width=.9\textwidth]{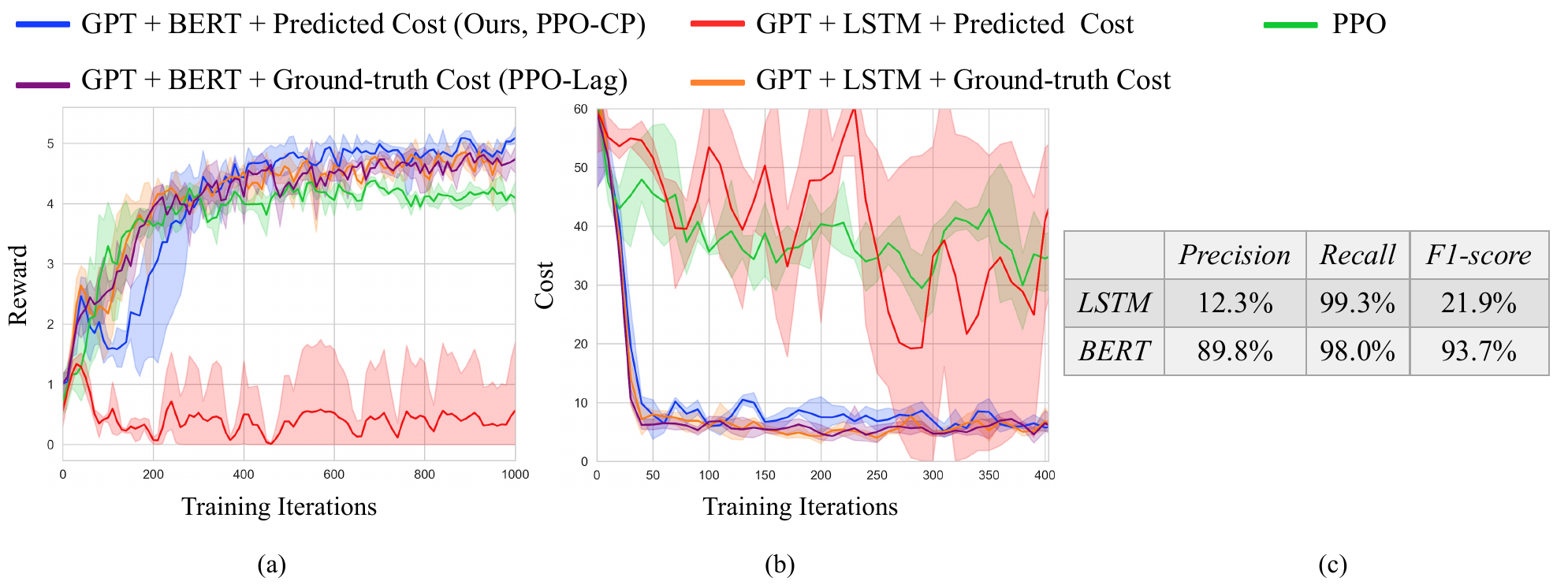}
    \caption{(a) gives the result of episode reward and (b) gives the result of episode cost as training goes on. The labels are in the format of '\textit{decoder model + encoder model + cost type}'. For example, '\textit{GPT + BERT + Predicted Cost}' stands for using GPT as decoder LM, BERT as encoder LM and predicted cost from our cost prediction module for policy learning. '\textit{GPT + BERT + Predicted Cost}' is our proposed method PPO-CP, and '\textit{GPT + BERT + Ground-truth Cost}' is PPO-Lag in Fig. \ref{fig:main_res}. (c) gives the cost prediction results when using BERT and LSTM as the encoder model. The cost prediction with LSTM as encoder has very poor performance, which leads to the collapsed training of \textit{'GPT + LSTM + Predicted Cost'} in (a) and (b).}
    \label{fig:ablate_bert}
\end{figure*}
General results on all five tasks are given in Fig. \ref{fig:main_res}. We report episode reward sum and episode cost sum, respectively to show how the methods perform. As we use cost prediction (CP) to replace ground-truth costs, we refer to our method as PPO-CP. 

From the results, we can see that compared with PPO that does not consider constraints, PPO-CP can learn safe policies like the baseline methods that require ground-truth costs. This indicates that our cost prediction module is very effective and helpful for learning safe policies. In some tasks, PPO-CP even learns safer policies than the baseline methods, which may be because the cost prediction module mistakenly thinks of some risky but potentially safe actions as violating the constraints (like walking on the edge of the cliff). Thus, the policy is trained to avoid these risky actions, resulting in fewer constraint violations and lower costs. PPO-CP and PPO-Lag achieve even higher rewards than PPO on \emph{Random} in Hazard-World-Grid. This is because they always tend to find the fastest way to obtain all the goal objects to finish the episode, as they will violate the constraints less if the episode ends fast. And the reward will decay within an episode. Therefore, PPO-CP and PPO-Lag can achieve even higher rewards than PPO.
\subsection{Ablation Study}
\subsubsection{Ablation on Contrastive Loss}
To see the effectiveness of fine-tuning BERT (encoder LM) for cost prediction, we run the proposed method and compare the cost prediction results on layout \emph{random} in Hazard-World-Grid with and without fine-tuning. The encoder LM is fine-tuned for 10 iterations. In each iteration, the encoder LM is trained with 128 randomly sampled condensed constraint pairs to minimize the contrastive loss in Eq. \ref{eq:contrast}.

The cost prediction results are given in Table \ref{tab:abla_contra}. From the results, we can see that our cost prediction module can predict costs accurately. Especially when the encoder LM is fine-tuned, the \textit{F1-score} of cost prediction is improved by 6.7\%. And it is worth noting that the \textit{Recall} (true positive rate) is improved by a large margin. As a result, the cost prediction module is much more sensitive to constraint violations, and the learned policy will follow the constraints better. 
\begin{table}[h]
        \vspace{-1em}
	\caption{Ablation study on contrastive loss in Eq. \ref{eq:contrast} to fine-tune encoder LM. Results show that contrastive loss can effectively improve cost prediction results, especially recall (true positive rate).}
	\label{tab:abla_contra}
	\begin{tabular}{cccc}\toprule
            & \textit{Precision} & \textit{Recall} & \textit{F1-score} \\ \midrule
            \textit{w} contrastive  & 89.8\%    & 98.0\% & 93.7\% \\
            \textit{w/o} contrastive & 87.4\%    & 86.6\% & 87.0\%  \\ \bottomrule
	\end{tabular}
\end{table}
\subsubsection{Ablation on Encoder LM}\begin{figure*}[t]
    \centering
    \includegraphics[width=.95\textwidth]{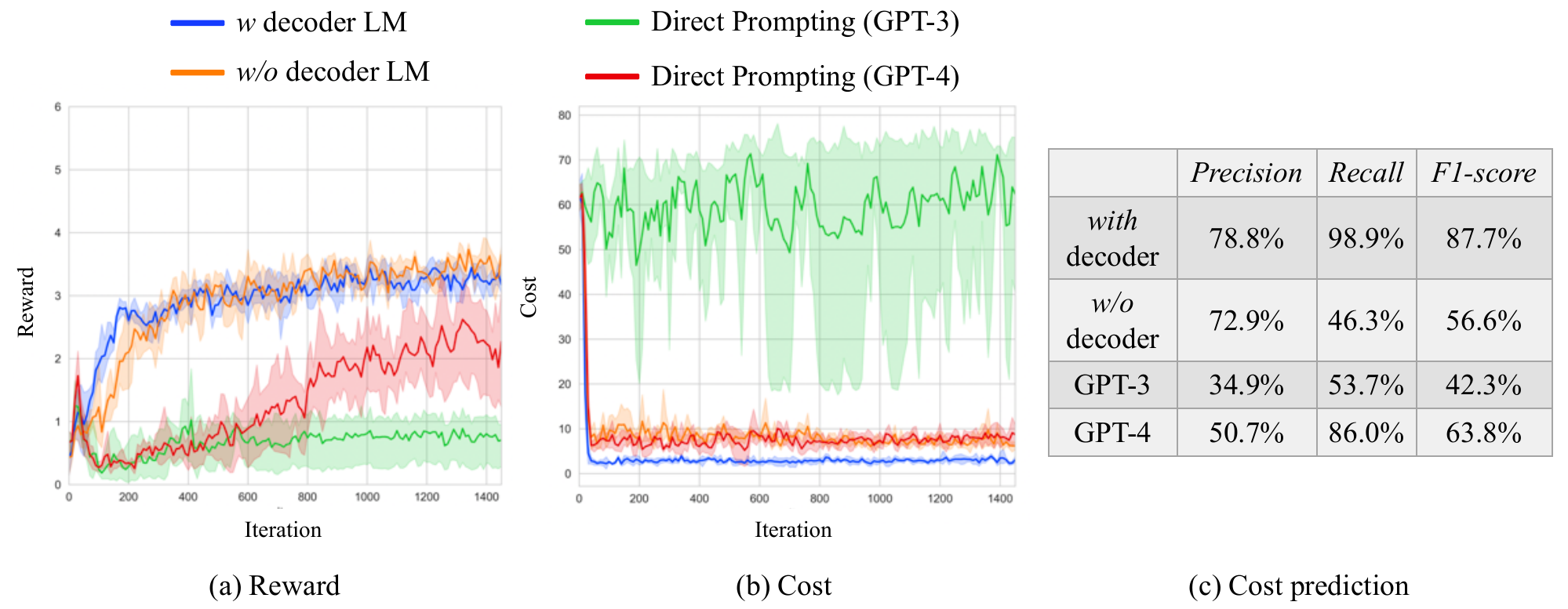}
    \caption{Experiment results for ablating decoder LM and directly prompting GPT for cost prediction. \textcolor{blue}{\emph{w} decoder LM} is our proposed method, \textcolor{orange}{\emph{w/o} decoder LM} removes the decoder LM and keeps the other modules such as cost prediction, \textcolor{green}{Direct Prompting GPT-3} removes our cost prediction module and query GPT-3 with the constraint and text-based observation for cost prediction, and \textcolor{red}{Direct Prompting GPT-4} replace GPT-3 with GPT-4 in Direct Prompting GPT-3. (a) gives the results of the episode reward. (b) is the results of episode cost and (c) gives the cost prediction results. The proposed method achieves the best performance on all three metrics while Directly prompting GPT-3 to perform the worst.}
    \label{fig:ablate_gpt}
\end{figure*}
BERT is pre-trained on large text corpora so that it can encode natural language input according to their semantic meaning. This is essential for our cost prediction because we need to compare the similarity between the natural language constraint and text-based observation. In this section, we ablate the encoder LM BERT from the proposed method and replace it with a long-short-term-memory (LSTM) network. We remain the other modules of the proposed method, such as GPT as decoder LM, and run an ablation study on layout \emph{random} in Hazard-World-Grid. As the randomly initialized LSTM cannot tell the difference among constraints, we also pre-train it with the contrastive loss for 10 iterations to make a fair comparison. The ablation results are given in Fig. \ref{fig:ablate_bert}. Labels are in the format of '\textit{decoder model + encoder model + cost type}'. For example, '\textit{GPT + BERT + Predicted Cost}' stands for using GPT as decoder LM, BERT as encoder LM and predicted cost from our cost prediction module for policy learning. {The decoder LM extracts semantic meaning from the constraints and is adopted in all methods except PPO, because it ignores the constraints. It is worth noting although PPO-Lag has access to the ground-truth cost, it still needs an Encoder LM to encode the natural language constraints, so that the constraints can be put into and constrain the policy network.}

From Fig. \ref{fig:ablate_bert}(c), we can see the cost prediction module performs poorly with LSTM as an encoder. Consequently, training of '\textit{GPT + LSTM + Predicted Cost}' collapses in Fig. \ref{fig:ablate_bert}(a) and (b). But when the ground-truth cost is available, '\textit{GPT + LSTM + Ground-truth Cost}' has very strong performance, which means the LSTM network is able to encode natural language constraint to constrain the policy, but unable to extract semantic meaning of constraints and observations to do cost prediction.
\subsubsection{Ablation on Decoder LM}
Our proposed method adopts GPT as a decoder LM to condense semantic meaning and eliminate ambiguity within the natural language constraints. To see the efficacy of the decoder LM, we ablate the decoder LM and use the encoder LM to directly encode original natural language constraints (such as the ones given in Fig. \ref{fig:envs}(b)). Experiments are run on layout \emph{longpath} in Hazard-World-Grid. We compare episode reward, episode cost and cost prediction results between the original method and the ablation.

From the results in Fig. \ref{fig:ablate_gpt}, we can see after ablating decoder LM from the cost prediction modules, our method (\emph{w} decoder LM) and \emph{w/o} decoder LM achieve similar episodes reward, but our method violates the constraints much less than the ablation. The average constraint violation for \emph{w} decoder LM is 3.2 per episode, while \emph{w/o} decoder LM violates the constraints 6.8 times per episode on average. And we can see from Fig. \ref{fig:ablate_gpt}(c) that the cost prediction results deteriorate dramatically, which demonstrates that it is necessary to adopt decoder LM to condense semantic meaning and eliminate ambiguity in the natural language constraints.
\subsubsection{Direct Prompting}
Through the acquisition of vast amounts of data, pre-trained large LMs, such as GPT \cite{gpt4}, Llama 2 \cite{touvron2023llama}, and PaLM 2 \cite{anil2023palm}, have demonstrated remarkable potential in achieving human-level intelligence. These models have strong capabilities in text generation tasks such as question answering. So, we directly prompt the large LMs with the text-based observation and natural language constraints and ask them to determine whether the constraints are violated. Then, we use the predicted costs by large LMs to train policies on layout \emph{longpath} and compare experiment results with the proposed method. The prompts we use are given in the supplementary material. The pre-trained models we consider in this study are GPT-3.5 \cite{brown2020language} and GPT-4 \cite{gpt4}, which are state-of-the-art pre-trained large LMs.

The results for direct prompting are given in Fig. \ref{fig:ablate_gpt}. Compared to other methods, the results show that Directly prompting GPT-3 performs poorly on all metrics, achieving the lowest reward and highest cost and struggling with cost prediction. As a successor of GPT-3, GPT-4 has a much stronger ability in causality. In our experiments, Directly prompting GPT-4 also results in better performance than GPT-3. With GPT-4 predicted costs, the policy can gradually learn to get high rewards, but it is still worse than using predicted cost by our cost prediction module. It is worth noting that directly prompting GPT-4 leads to a large improvement in the Recall rate, which means the number of False Negative predictions is dramatically reduced. As a result, directly prompting GPT-4 can also learn a safe policy compared to directly prompting GPT-3.
\section{Conclusion and Future Work}
In this paper, we study the problem of safe RL with free-form natural language constraints and propose to use pre-trained LMs for cost prediction. A decoder LM and an encoder LM are adopted to process, comprehend and encode the constraints, which is later used to predict costs. Compared to previous methods dealing with safe RL with natural language constraints, our proposed method does not require ground-truth costs, can handle much more free-form constraints and does not need to learn an extra module to model entities within the environments. Empirically, we evaluate the proposed method on grid-world navigation tasks and robot control tasks against baseline methods that require ground-truth costs. Experiment results demonstrate that our method can accurately predict costs and learn safe policies that enable agents to accomplish given tasks while obeying the constraints. Extensive ablation studies are also conducted to show the efficacy of each module in our method.

In the future, besides learning safe policies with pre-trained LMs, we will also explore how to make agents' decisions more interpretable through pre-trained LMs, which is crucial for safety-critical tasks such as autonomous driving.





\balance
\bibliographystyle{ACM-Reference-Format} 
\bibliography{sample}


\end{document}